\documentclass[twocolumn,letterpaper]{IEEEAerospaceCLS}  % only supports two-column, letterpaper format

\usepackage[]{graphicx,float,latexsym,amssymb,amsfonts,amsmath,times,epsfig,url,multirow}
\newcommand{\ignore}[1]{}  % {} empty inside = %% comment

\begin{document}
\title{Self-Attending Task Generative Adversarial Network for Realistic Satellite Image Creation}

\author{%
Nathan Toner\\ 
KBR\\
985 Space Center Dr.\\
Colorado Springs, CO 80915\\
nathan.toner@us.kbr.com
\and 
Justin Fletcher\\
Air Force Research Laboratory\\
550 Lipoa Parkway\\
Kihei, HI 96753\\
justin.fletcher.7@us.af.mil
%%%% IMPORTANT: Use the correct copyright information--IEEE, Crown, or U.S. government. %%%%%
% \thanks{\footnotesize 978-1-6654-3760-8/22/$\$31.00$ \copyright2022 IEEE}              % This creates the copyright info that is the correct 2021 data.
%\thanks{{U.S. Government work not protected by U.S. copyright}}         % Use this copyright notice only if you are employed by the U.S. Government.
%\thanks{{978-1-6654-3760-8/22/$\$31.00$ \copyright2022 Crown}}          % Use this copyright notice only if you are employed by a crown government (e.g., Canada, UK, Australia).
%\thanks{{978-1-6654-3760-8/22/$\$31.00$ \copyright2022 European Union}}    % Use this copyright notice is you are employed by the European Union.
}

\maketitle

\thispagestyle{plain}
\pagestyle{plain}

\maketitle

\thispagestyle{plain}
\pagestyle{plain}

\begin{abstract}
We introduce a self-attending task generative adversarial network (SATGAN) and apply it to the problem of augmenting synthetic high contrast scientific imagery of resident space objects with realistic noise patterns and sensor characteristics learned from collected data. Augmenting these synthetic data is challenging due to the highly localized nature of semantic content in the data that must be preserved. Real collected images are used to train a network what a given class of sensor's images should look like. The trained network then acts as a filter on noiseless context images and outputs realistic-looking fakes with semantic content unaltered. The architecture is inspired by conditional GANs but is modified to include a task network that preserves semantic information through augmentation. Additionally, the architecture is shown to reduce instances of hallucinatory objects or obfuscation of semantic content in context images representing space observation scenes.
\end{abstract}

% \tableofcontents

%%%%%%%%%%%%%%%%%%%%%%%%%%%%%%%%%%%%%%
\section{Introduction}
\begin{figure*}[!htb]
  \centering
  % \begin{minipage}[b]{0.32\linewidth}
  %   \centering
  %   \centerline{\includegraphics[width=\textwidth]{Figures/task_b1.png}}
  % \end{minipage}
  \begin{minipage}[b]{0.49\linewidth}
    \centering
    \centerline{\includegraphics[width=\textwidth]{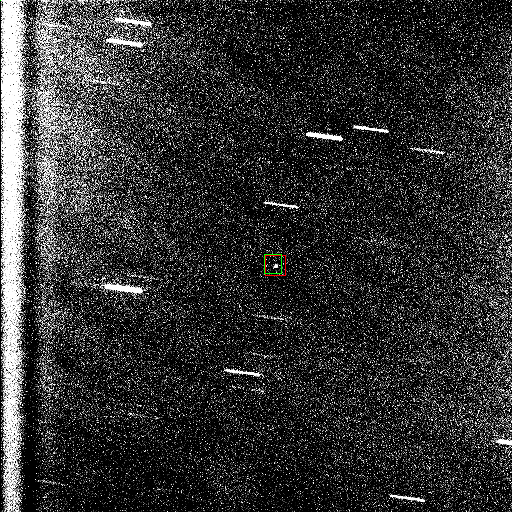}}
  \end{minipage}
  \begin{minipage}[b]{0.49\linewidth}
    \centering
    \centerline{\includegraphics[width=\textwidth]{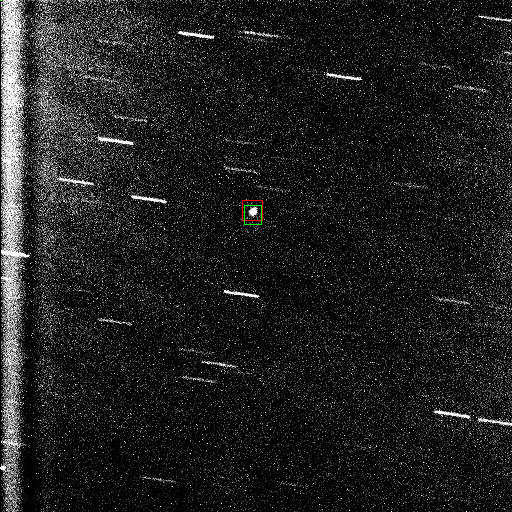}}
  \end{minipage}
  \caption{\textbf{Sample electro-optical images containing GEO objects including annotated true (red) and predicted (green)
    bounding boxes. Images of this kind represent the target domain of SATGAN.}}
  \label{fig:task_b}
\end{figure*}

%%%%%%%%%%%%%%%%%%%%%%%%%%%%%%%%%%%%%%
\label{sec:intro}
Successful tracking and management of artificial Earth satellites in geosynchronous Earth orbit (GEO) requires that they first be detected. These objects---commonly called resident space objects (RSOs)---are typically observed by electro-optical (EO) telescopes, which capture high contrast scientific imagery (see Figure~\ref{fig:task_b}) in which GEO RSOs appear as point sources. This imagery is processed to extract the location of objects. Fletcher, \emph{et al.}\ \cite{fletcher_feature-based_2019,fletcher_dynamical_2021} is the current state-of-the-art in this object detection task for single frames, and employs a deep learning method based on the ubiquitous ``you only look once'' (YOLO) architecture \cite{redmon_you_2015,redmon_yolo9000:_2016,redmon_yolov3:_2018}. Like many deep learning approaches to computer vision tasks, this method requires representative data from the domain in which the task is to be performed. A variety of EO sensors, each having different fundamental properties, are used for space object detection and new sensors are developed regularly. However, current EO datasets contain only a small subset of existing sensor technology and may not represent future sensors well. Furthermore, collected data under-represents rare but important events such as collisions and breakups. Synthetic images of EO scenes are used to model new sensors and rare events, but models trained on this data generalize poorly to real images \cite{fletcher_feature-based_2019}. This generalization gap is caused in part by noise that is present in the real data, but is impractical to model precisely, such as stray light, cloud scatter, and mount effects. Thus, new approaches are needed to augment synthetic data generation for training computer vision models for RSO detection.

In this work, we explore the use of generative adversarial networks (GANs) to generate realistic imagery of scenes comprising RSOs in GEO observed using EO telescopes. An existing dataset known as SatNet \cite{fletcher_feature-based_2019}---comprising EO observations from half-meter-class telescopes \cite{sabol_recent_2002}---is used as the target distribution of the proposed GAN architecture. A generator model is conditioned on noiseless contextual images of arbitrary scenes created with the SatSim space scene simulator and trained to produce realistic noisy image that retain semantic information, i.e., RSOs in the scene. We contribute 1) a novel GAN architecture that incorporates task-consistency and additive noise that may be used to enhance synthetic data generation for scientific images and 2) an analysis of sim2real generalization performance of this architecture. 

We propose a method for generating realistic additive sensor noise using a self-attending task GAN (SATGAN) based loosely on the conditional GAN (CGAN) ``pix2pix'' architecture developed by Isola \emph{et al.}\ \cite{isola_image--image_2018}. Background information and similar applications are discussed in Section~\ref{sec:related}. The baseline pix2pix model used in this study is discussed in
Section~\ref{sec:pix2pix}. An overview of the SATGAN architecture is given in Section~\ref{sec:semi-cond-task}. Experimental procedures and results comparing the performance of the YOLO-based satellite detection network of \cite{fletcher_feature-based_2019} on real, simulated, and generated data are discussed in Section~\ref{sec:results}. Finally conclusions and future work are discussed in Section~\ref{sec:conclusions_and_future}.

%%%%%%%%%%%%%%%%%%%%%%%%%%%%%%%%%%%%%%
\section{Related work}
%%%%%%%%%%%%%%%%%%%%%%%%%%%%%%%%%%%%%%
\label{sec:related}

Data-driven methods enable superior perceptual automation for deep space object detection in EO imagery. For example, in \cite{fletcher_feature-based_2019}, a specialized model based on the YOLO network is trained to outperform the state of the art in detection of dim and closely spaced objects in space scenes. As with most deep learning models, this approach requires significant volumes of data. Datasets such as SatNet exist, but data available from past sensors may not be representative of the future and collecting and annotating data is a costly and time-consuming process.

We have the ability to simulate the geometry of future space events using a high-fidelity simulator tool called SatSim, but realistic sensor noise is difficult to simulate. Methods such as the YOLO-based approach mentioned above have been shown to perform very well on data drawn from the same domain (i.e., sensor type) as the training dataset, but performance degrades when these algorithms are required to perform across domains, requiring retraining of the neural network on potentially large datasets that may be difficult to obtain. The goal of this work is to create a model that maps synthetic space scenes from simulation along with semantic information about object location to a filtered image, such that the resulting filtered images are indistinguishable from real images taken by target sensors while maintaining the semantic information present in the simulated scene.

In their seminal work, Shrivastava \emph{et al.}\ \cite{shrivastava_learning_2017} developed a CGAN conditioned on artificial images of eyes with gaze direction semantic information that produced realistic-looking eye images while maintaining this contextual information. They introduced a novel self-regularization term that minimized the pixel-to-pixel $l_1$ norm in a learned feature space between the contextual input image and the resulting filtered image. This loss, along with other best practices, was shown to maintain semantic information from the prior while allowing the generator to produce realistic-looking images.

The pix2pix architecture proposed by Isola \emph{et al.}\ \cite{isola_image--image_2018} uses a CGAN to learn general-purpose image-to-image translation tasks. This architecture conditions both the generator \emph{and} discriminator on an image drawn from an input domain, and not only learns the mapping from input domain to target domain, but also learns a loss function to train this mapping, making it applicable to a wide variety of problems. A pix2pix model was trained to act as a baseline for this study, but was omitted from the analysis of detection performance due to its poor resulting image quality (see Figure~\ref{fig:pix2pix-res}).

Several additional approaches to image-to-image translation have been proposed including Zhu \emph{et al.}'s CycleGAN \cite{zhu_unpaired_2018}, which builds on the pix2pix network by introducing a cycle consistency loss and an encoder--decoder framework that allows images to be transferred from both the input domain to the target domain and from the target domain to the input domain. This allows the network to be trained such that images that are semantically similar to one another will remain similar to one another across domains. This cycle consistency loss acts as a regularizer that ensures semantic information is maintained in an unsupervised manner through transformations in both directions (encoding and decoding).

Ouyang \emph{et al.}\ introduced Task GAN \cite{ouyang_task-gan_2019} to improve the ability of a GAN to maintain semantic information that may be present in only a small portion of an input image. In \cite{ouyang_task-gan_2019}, the authors train a GAN for image restoration, using low- and high-resolution images as input and target domain respectively, but also train a task network alongside this GAN designed to perform a semantically meaningful task on real and filtered images. They demonstrate effective image restoration for pathology slides in which small, highly localized, and detailed features are present and important.

%%%%%%%%%%%%%%%%%%%%%%%%%%%%%%%%%%%%%%
\section{Pix2pix}
%%%%%%%%%%%%%%%%%%%%%%%%%%%%%%%%%%%%%%
\label{sec:pix2pix}

Isola \emph{et al.}\ \cite{isola_image--image_2018} developed the pix2pix CGAN architecture that provides fully supervised training conditioned on some semantic context input for the purposes of bridging data domains. In this architecture, given an input image $x$ and a contextual image plus noise $z = c + w$ with $w\sim {\mathcal{N}}(\mu_w,\, \sigma_w^2)$, a discriminator $D(x,\, c;\, \theta_D)$ parameterized by $\theta_D$ \emph{and conditioned on $c$} is trained to classify inputs as either real or fake images and a generator $\hat{x} = G(z;\, \theta_G)$ parameterized by $\theta_G$ is trained such that the resulting fake images $\hat{x}$ are indistinguishable from real images $x$. The training process is an adversarial optimization process in which $\theta_G$ and $\theta_D$ are optimized iteratively and in turn such that a Nash equilibrium is located according to (\ref{eqn:gan-obj}), where ${\mathbb{E}}[\cdot]$ is the expected value of its argument \cite{farnia_gans_2020}.

\begin{equation}
  \label{eqn:gan-obj}
  \min_{\theta_G} \max_{\theta_D} {\mathbb{E}}_x\big[\log D(x,\, c)\big] + {\mathbb{E}}_z\big[ \log 1 - D\big(G(z),\, c \big)\big]
\end{equation}

The pix2pix architecture utilizes a GAN training paradigm with a generator seeking to minimize a loss ${\mathcal{L}}_G$ comprising an adversarial loss and a reproduction loss; and a discriminator seeking to minimize ${\mathcal{L}}_D$, the cross entropy between expected and predicted input labels (real or fake). The total loss is shown in (\ref{eqn:pix-loss}), where the relative weights of loss components are controlled by the hyperparameters $\alpha$ and $\lambda$.

\begin{align}
\label{eqn:pix-loss}
  {\mathcal{L}}_G &= \alpha {\mathbb{E}}_{x, \hat{x}}\lVert \hat{x} - x \rVert_1 - {\mathbb{E}}_{\hat{x}}\big[\log D(\hat{x}) \big] \notag \\
  {\mathcal{L}}_D &= -{\mathbb{E}}_x\big[\log D(x) \big] - {\mathbb{E}}_{\hat{x}}\big[\log\big(1 - D(\hat{x}) \big) \big] \notag \\
  {\mathcal{L}} &= {\mathcal{L}}_G + \lambda {\mathcal{L}}_D
\end{align}

\begin{figure*}[!htb]
  \centering
  \includegraphics[width=\textwidth]{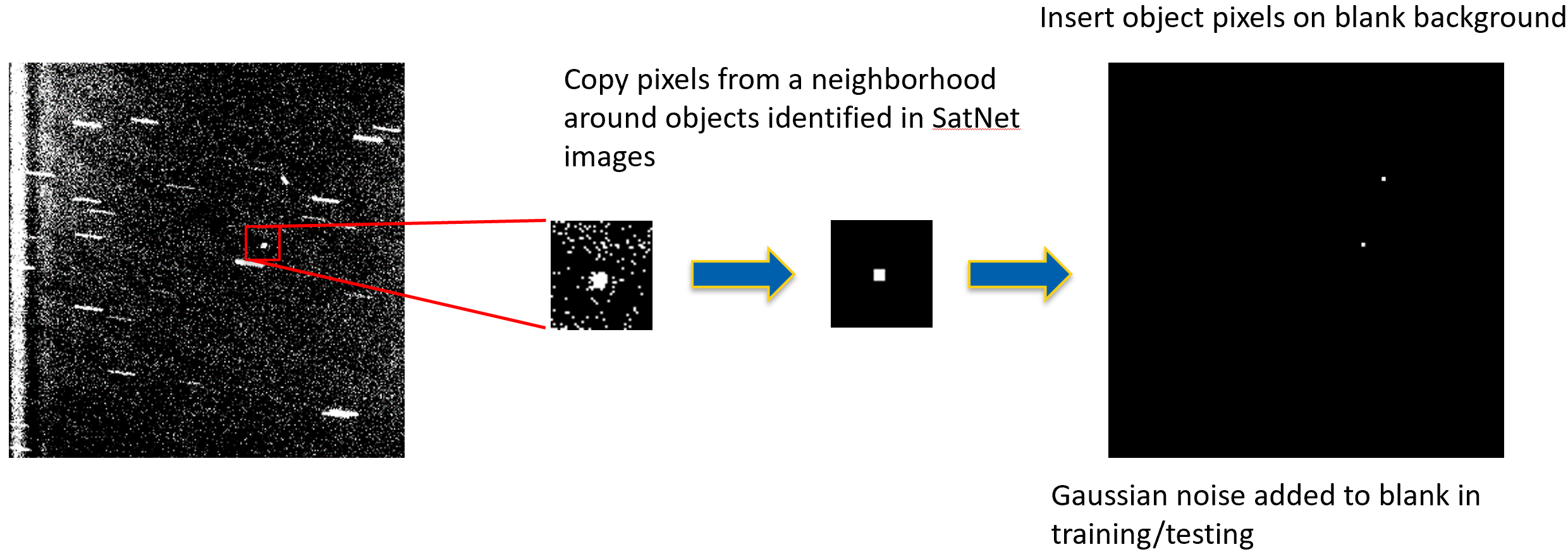}
  \caption{\textbf{Context image generation from labeled target SatNet images.}}
  \label{fig:image-blanks}
\end{figure*}

\begin{figure*}[!htb]
  \centering
  \includegraphics[width=\textwidth]{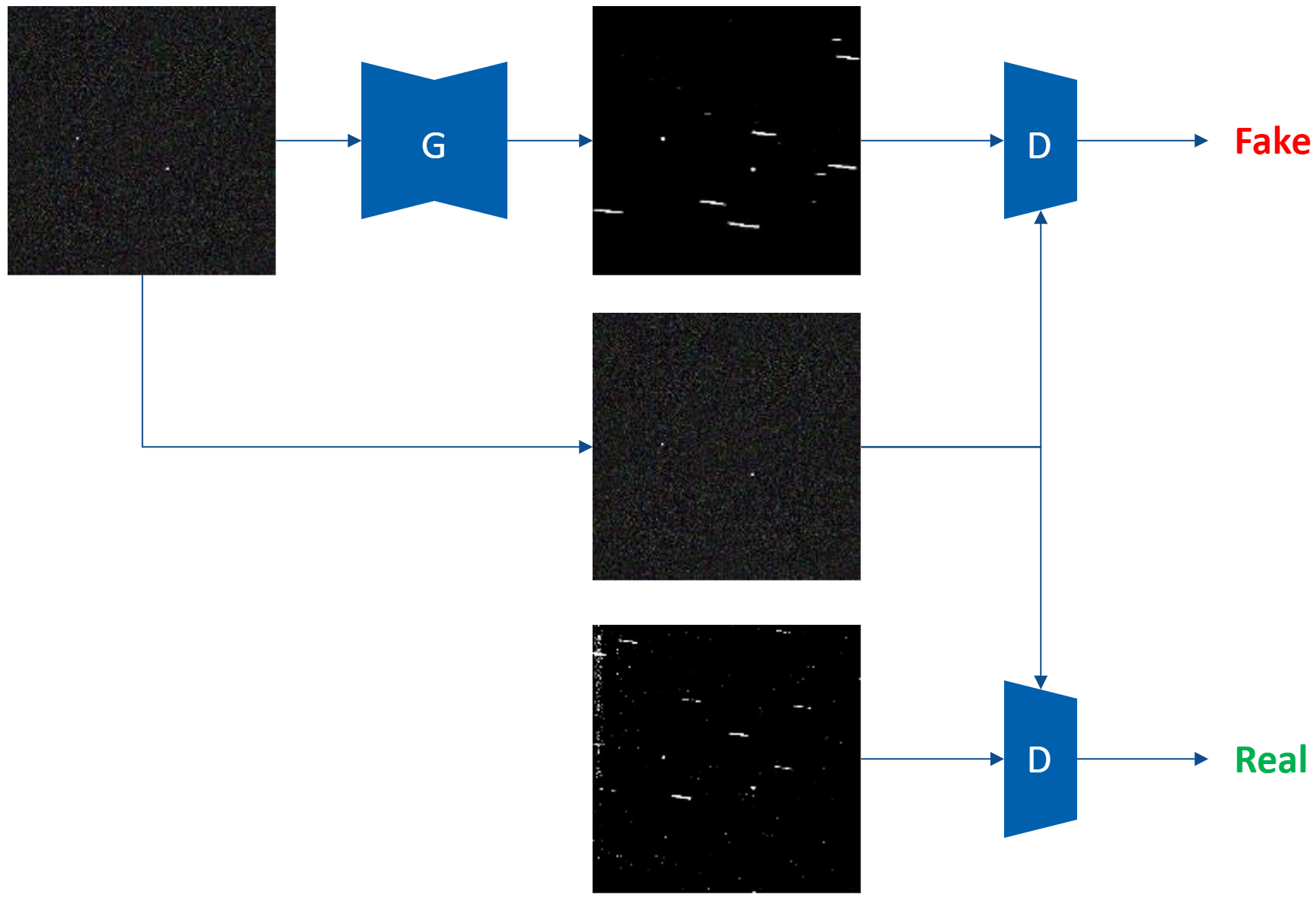}
  \caption{\textbf{Pix2pix model architecture diagram.}}
  \label{fig:pix2pix-arc}
\end{figure*}

A pix2pix architecture was trained as a baseline for this study, with ``blank'' context images generated from labeled target images. These context images were created by copying the pixels in a region around labeled objects onto a uniform image of the average intensity of images in the training set. This process is illustrated in Figure~\ref{fig:image-blanks}. The reliance of pix2pix on these context images makes a strict requirement for a large corpus of labeled target domain data. Eliminating this requirement was a key motivator behind the design of the SATGAN architecture. The pix2pix architecture is illustrated in Figure~\ref{fig:pix2pix-arc} where the dependence of the discriminator on both context and sample images is made clear. Pix2pix network performance is discussed in the Results section.

%%%%%%%%%%%%%%%%%%%%%%%%%%%%%%%%%%%%%%
\section{Self-attending task GAN}
%%%%%%%%%%%%%%%%%%%%%%%%%%%%%%%%%%%%%%
\label{sec:semi-cond-task}

\begin{figure*}[!htb]
  \centering
  \includegraphics[width=\textwidth]{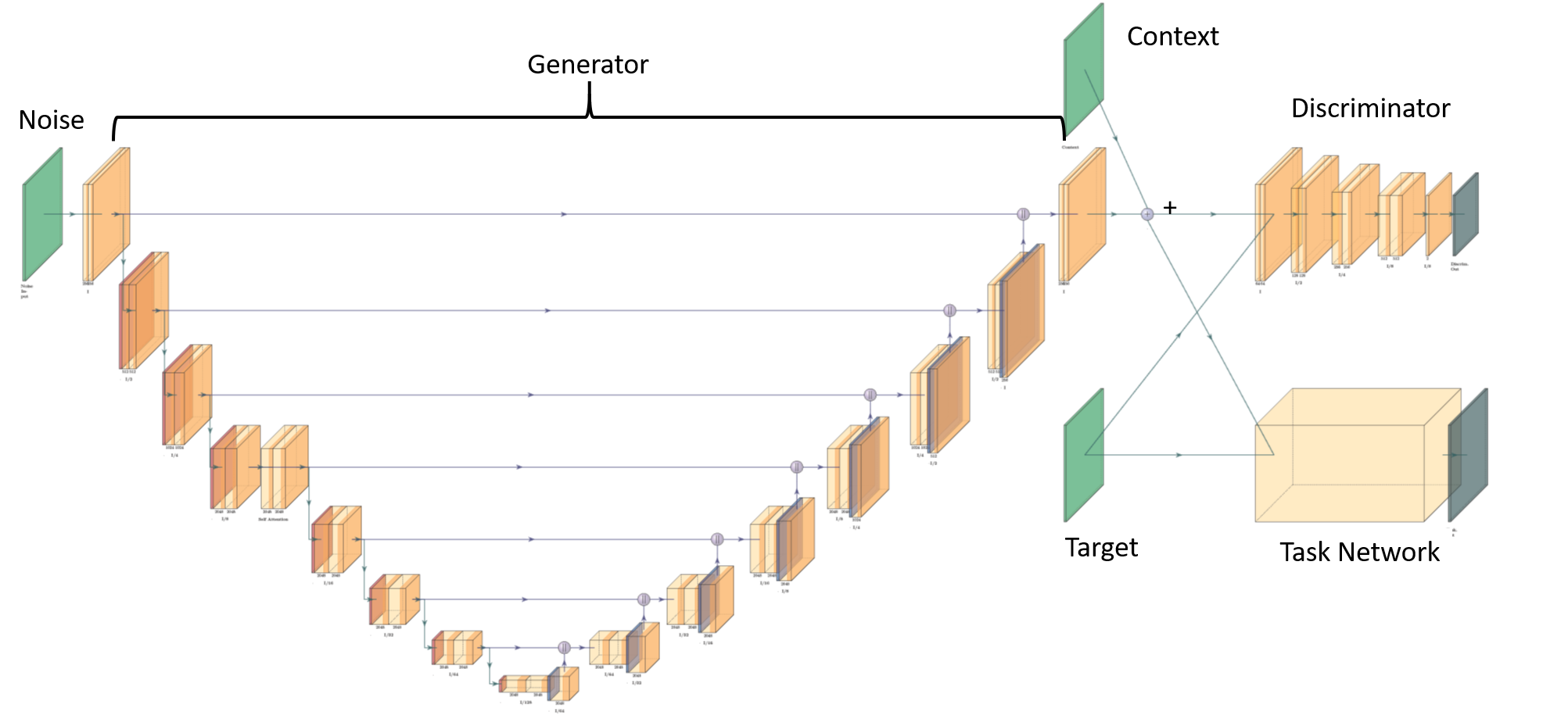}
  \caption{\textbf{Self-attending task GAN architecture comprising U-net-based generator,
  patch-GAN-based discriminator, and YOLO-based task network.}}
  \label{fig:semi-cond-task}
\end{figure*}

Starting from pix2pix, the SATGAN architecture was developed to remove the dependency between target and context images---allowing for successful training with far fewer target images and no requirement for labeled target images---and, through the addition of the task network, to maintain semantic information through the generator. The SATGAN architecture shown in Figure~\ref{fig:semi-cond-task} takes as input a noise field $z \sim {\mathcal{N}}(\mu_z, \sigma_z^2)$ and any simulated, \emph{noiseless} scene as context $c$, as well as real images from the target sensor $x$ with or without labeled RSOs. The change to noiseless context images means that any arbitrary simulated scene of space objects can be used to produce any arbitrary space scene with generated sensor noise. This enables the trained generator to be used to create limitless artificial data to augment real training data.

The generator learns to emulate the target sensor's characteristics to produce realistic fake sensor noise given Gaussian noise input $\tilde{n} = G(z)$. This generated sensor noise is then added to the original noiseless context image to create realistic fake images $\hat{x} = c + \tilde{n}$ to fool a discriminator network $D(x)$. Note that the SATGAN discriminator no longer depends directly on the context image $c$. A separate task network $T(x;\, \theta_T)$ parameterized by $\theta_T$ is pre-trained to perform a context-specific task on the images or is trained \emph{in situ} on simulated input images and (where available) real images. The task network takes as inputs real or fake images $x$ or $\hat{x}$ and returns estimated labels for those images $\hat{y} | x, \hat{x}$. This task network has two effects on the resulting architecture:

\begin{enumerate}
\item It tends to prefer generators that do not alter the semantic content of images, i.e., that do not add hallucinated objects or remove or obscure existing objects.
\item It acts as a measure of the convergence of domains: as the task performance on GAN-filtered and target images converges, the domains are likely to converge.
\end{enumerate}

The SATGAN architecture utilizes a task-GAN training paradigm with a generator and discriminator similar to pix2pix using the same losses defined in (\ref{eqn:pix-loss}). A task-specific loss ${\mathcal{L}}_T$ is added comprising a task-specific loss function $f_T(x,\, y)$ that operates on real or fake images $x$ or $\hat{x}$ and their associated task-specific targets $y$ or $y_c$, where $y_c$ are the targets of context image $c$. Task loss and total total loss are shown in (\ref{eqn:task-loss}) where ${\mathcal{L}}_G$ and ${\mathcal{L}}_D$ are defined in (\ref{eqn:pix-loss}); the relative weights of the loss components are controlled by the hyperparameters $\alpha$, $\beta$, $\lambda$, and $\gamma$; and the YOLO loss was used for $f_T$ \cite{redmon_you_2015,redmon_yolo9000:_2016,redmon_yolov3:_2018}.

\begin{align}
\label{eqn:task-loss}
  {\mathcal{L}}_T &= {\mathbb{E}}_{x, \hat{x}} \big[f_T\big(T(x),\, y\big) + \beta f_T\big(T(\hat{x}),\, y_c \big)\big] \notag \\
  {\mathcal{L}} &= {\mathcal{L}}_G + \lambda {\mathcal{L}}_D + \gamma {\mathcal{L}}_T
\end{align}

% TODO (NLT): maybe discuss the resulting architecture in statistical terms.
% \hat{x} conditioned on c and so forth.

The SATGAN architecture is able to learn target sensor characteristics with or without labeled target data, and with a much smaller corpus of target sensor data than that required for previous conditional and semi-conditional GAN architectures like pix2pix and CycleGAN. It is important to note that the $l_1$ reproduction loss is used in the generator. This loss tended to result in hallucinatory objects when using conditional or semi-conditional GAN (i.e., pix2pix) for the space object detection problem investigated herein. The task network played a necessary role in preventing the generator from learning to hallucinate objects.

For all experiments, the generator chosen was based on the U-net architecture used in \cite{isola_image--image_2018} with a self-attention layer added after the third encoder layer. This self-attention layer appeared to qualitatively improve the realism of generated sensor noise. The discriminator was based on the PatchNet used in \cite{isola_image--image_2018}. The task network was the SatNet YOLO network presented in \cite{fletcher_feature-based_2019}. All architectures were written in TensorFlow using the 2.2 Python API and trained on a single Nvidia~V100 GPU with 32~GB of memory. Code is available at \url{https://github.com/Engineero/satgan}.

% TODO (NLT): public release of github code? 

%%%%%%%%%%%%%%%%%%%%%%%%%%%%%%%%%%%%%%
\section{Results}
%%%%%%%%%%%%%%%%%%%%%%%%%%%%%%%%%%%%%%
\label{sec:results}

Task network results on sample target domain (real) images with bounding boxes drawn around true (red) and predicted (green) objects are shown in Figure~\ref{fig:task_b}. Note that bounding boxes were predicted with an implementation of the SatNet YOLO architecture presented in \cite{fletcher_feature-based_2019}.

\begin{figure*}[!htb]
  \centering
  \begin{minipage}[b]{0.49\linewidth}
    \centering
    \centerline{\includegraphics[width=\textwidth]{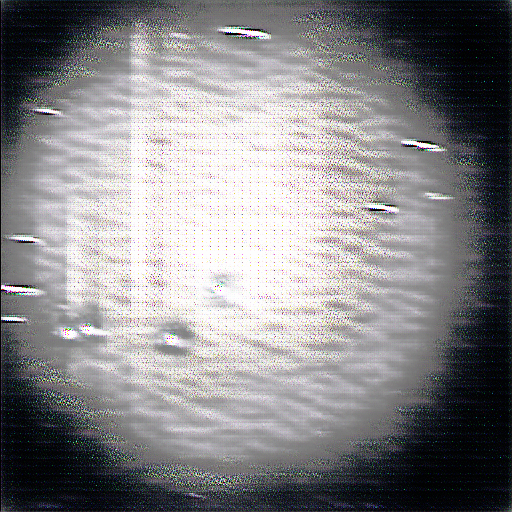}}
  \end{minipage}
  \begin{minipage}[b]{0.49\linewidth}
    \centering
    \centerline{\includegraphics[width=\textwidth]{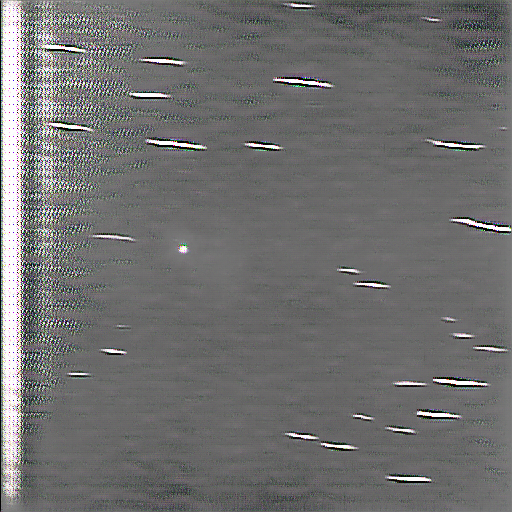}}
  \end{minipage}
  \caption{\textbf{Sample results from pix2pix conditional GAN architecture.}}
  \label{fig:pix2pix-res}
\end{figure*}

A baseline pix2pix architecture was trained using ``blank'' images as context built by taking pixels associated with objects from labeled target domain data and copying them onto an otherwise uniform image (see Figure~\ref{fig:image-blanks}). Note that this approach requires a large corpus of \emph{labeled} target domain data. Resulting generated images from this architecture are shown in Figure~\ref{fig:pix2pix-res}. Comparing these images to the sample target images shown in Figure~\ref{fig:task_b}, we see that the pix2pix architecture learns some sensor characteristics, but fails to emulate higher-fidelity noise characteristics. It was also found that semantic content in images as measured by task network performance was often hallucinated or destroyed by the pix2pix architecture.

\begin{figure*}[!htb]
  \centering
  % \begin{minipage}[b]{0.32\linewidth}
  %   \centering
  %   \centerline{\includegraphics[width=\textwidth]{Figures/input_a1.png}}
  % \end{minipage}
  \begin{minipage}[b]{0.49\linewidth}
    \centering
    \centerline{\includegraphics[width=\textwidth]{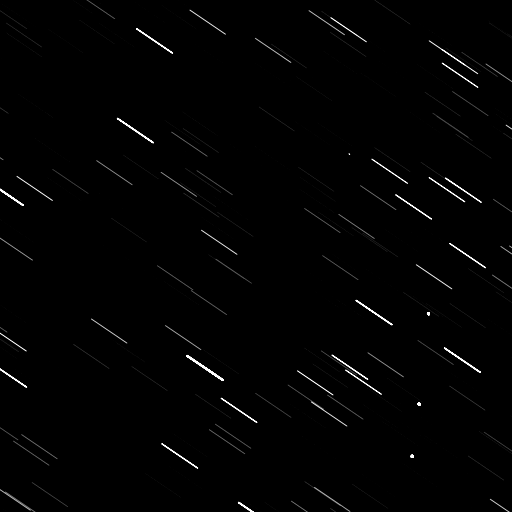}}
  \end{minipage}
  \begin{minipage}[b]{0.49\linewidth}
    \centering
    \centerline{\includegraphics[width=\textwidth]{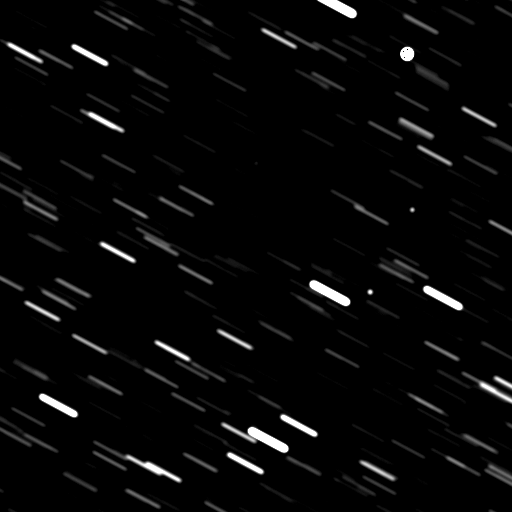}}
  \end{minipage}
  \caption{\textbf{Sample contextual inputs to Task-GAN generator network.}}
  \label{fig:inputs}
\end{figure*}

The SATGAN architecture learns to generate realistic sensor characteristics and adds these to arbitrary simulated, noiseless scenes created using SatSim, a high-fidelity simulation tool. Sample noiseless context images generated with SatSim are shown in Figure~\ref{fig:inputs}.

\begin{figure*}[!htb]
  \centering
  % \begin{minipage}[b]{0.32\linewidth}
  %   \centering
  %   \centerline{\includegraphics[width=\textwidth]{Figures/task_a1.png}}
  % \end{minipage}
  \begin{minipage}[b]{0.49\linewidth}
    \centering
    \centerline{\includegraphics[width=\textwidth]{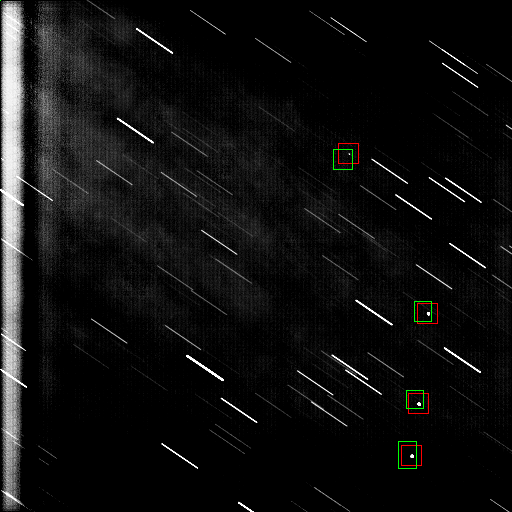}}
  \end{minipage}
  \begin{minipage}[b]{0.49\linewidth}
    \centering
    \centerline{\includegraphics[width=\textwidth]{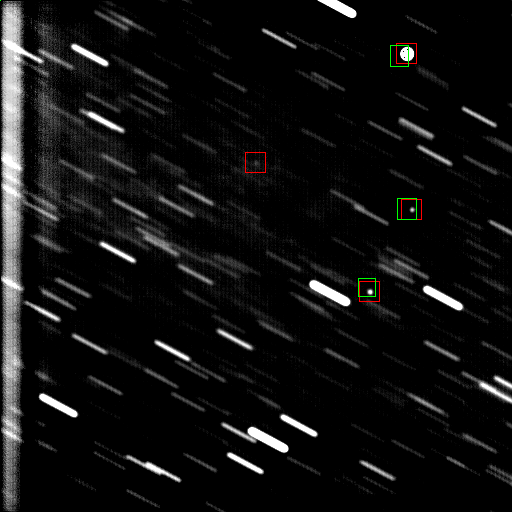}}
  \end{minipage}
  \caption{\textbf{Resulting outputs from SATGAN generator with true (red) and
    predicted (green) bounding boxes.}}
  \label{fig:task_a}
\end{figure*}

Resulting simulated images with bounding boxes drawn around true (red) and predicted (green) objects are shown in Figure~\ref{fig:task_a}. Predicted bounding boxes were drawn using the same implementation of SatNet YOLO as in Figure~\ref{fig:task_b}. Comparing Figure~\ref{fig:task_a} with Figure~\ref{fig:task_b}, we observe qualitatively that the generated noise characteristics are very similar to the true sensor characteristics, including large-scale structured noise, shot noise, and hot and dead pixels. Furthermore, the task network detections shown in both figures (green boxes) indicate that the task network performs similarly across the domains, showing qualitatively a degree of domain transfer. Note that the simulated data includes dimmer, more difficult-to-detect objects than the target domain, resulting in lower overall performance on this domain despite realistic generated noise characteristics. This is indeed one of the advantages of improving synthetic image quality: difficult-to-detect objects can easily be incorporated into training for future detection networks.

\begin{figure*}[!htb]
    \centering
    \includegraphics[width=.49\linewidth]{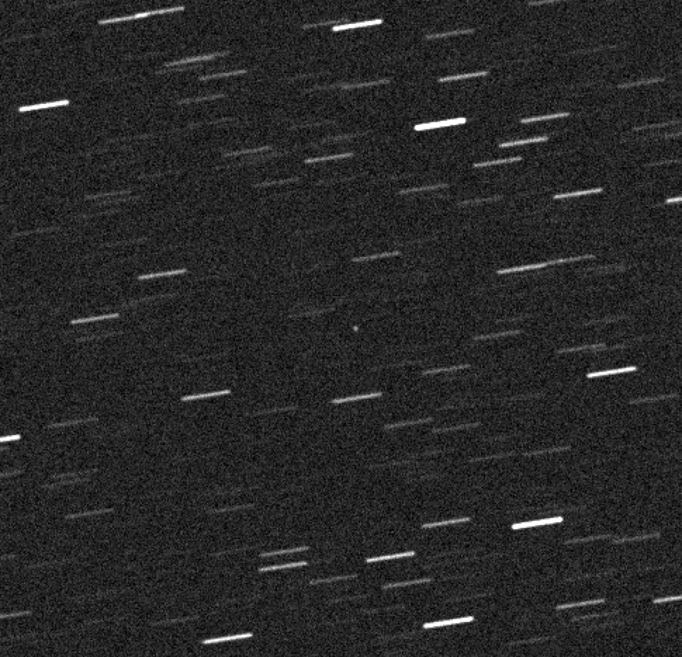}
    \caption{\textbf{Simulated image from SatSim.}}
    \label{fig:satsim}
\end{figure*}

For comparison, a simulated image with \emph{physics-based} simulated noise is shown in Figure~\ref{fig:satsim}. This image represent physics-based simulation of sensor properties emulating the target domain images shown in Figure~\ref{fig:task_b}. By comparing this fully-simulated image to the simulated images with SATGAN-\emph{generated} noise, we can see that SATGAN generates images that are qualitatively more convincing fakes of the target domain. The generator has learned to create noise characteristics that are difficult to model, such as  image-wide structured noise patterns and hot pixels. 

\begin{figure*}[!htb]
    \centering
    \includegraphics[width=0.75\textwidth]{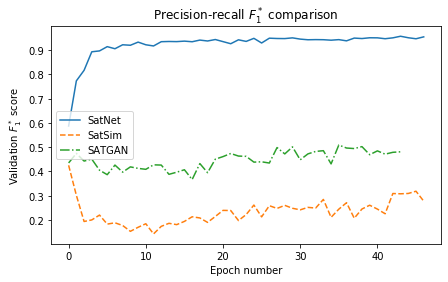}
    \caption{\textbf{Max F1 scores for networks trained on SatNet (target domain),
    SatSim-simulated, and SATGAN-generated data and validated on SatNet data.}}
    \label{fig:max_f1}
\end{figure*}

For quantitative comparison, a new YOLO SatNet network was trained on target domain data, simulated data using SatSim, and generated data using the SATGAN generator to produce realistic sensor noise. Performance metrics between these three networks were compared on the \emph{target} domain validation set, including the maximum F1 ($F_1^*$) score of each network's precision-recall curve. These $F_1^*$ scores on the target domain validation set are shown for the first 45 epochs of training in Figure~\ref{fig:max_f1}. The legend in the figure indicates on which dataset each network was trained.

Figure~\ref{fig:max_f1} shows a clear performance improvement for a network trained on SATGAN-generated data over a network trained on SatSim-simulated data when evaluated on target domain data. Other than a spurious first epoch in which the SatSim-trained network performs unusually well, the SATGAN-trained network typically outperforms it by a factor of two. While the SATGAN-trained YOLO network is not as performant on the target domain as a network trained on the target domain directly, this result indicates an improvement over state-of-the-art simulation for bridging the gap between simulated and real data domains. Details of the best-performing epoch for each network are given in Table~\ref{tab:perf-compare}. Note that for the network trained on SatSim data, the best-performing epoch was the first as can be seen in Figure~\ref{fig:max_f1}. This behavior merits further investigation.

\begin{table}[!htb]
    \begin{center}
    % \vspace{-0.4cm}
    \caption{\textbf{Best-epoch F-measure performance comparison on SatNet validation dataset.}\label{tab:perf-compare}}
    \vspace{0.2cm}
    \begin{tabular}{l|ccc}
        \centering \multirow{2}{*}{\textbf{Training Dataset}}& \multicolumn{3}{c}{\textbf{SatNet Validation Results}} \\

         & Precision & Recall & $F_1^*$ \\
         
        \hline
        \hline 
        SatNet (target) & 0.968 & 0.948 & 0.958 \\
        \hline
        SatSim (baseline) & 0.507 & 0.371 & 0.429 \\
        SatSim+SATGAN & \textbf{0.552} & \textbf{0.471} & \textbf{0.509} \\
    \end{tabular}
    %\vspace{-0.5cm}
    \end{center}
\end{table}

\begin{figure}[!htb]
    \centering
    \includegraphics[width=\linewidth]{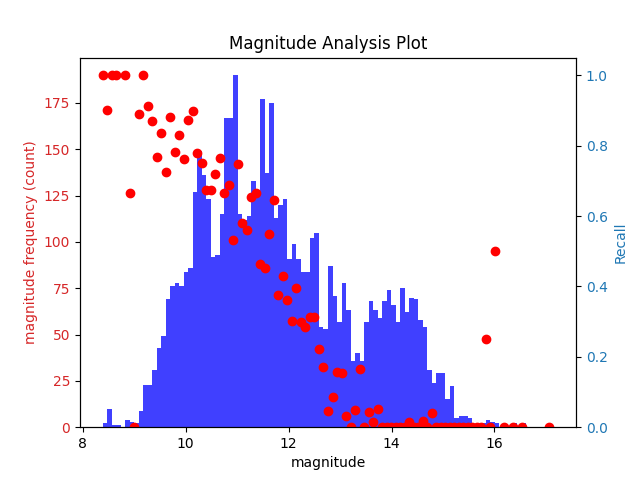}
    \caption{\textbf{Recall vs.\ visual magnitude of network trained on \textit{simulated} images.}}
    \label{fig:mag-recall-sim}
\end{figure}

\begin{figure}[!htb]
    \centering
    \includegraphics[width=\linewidth]{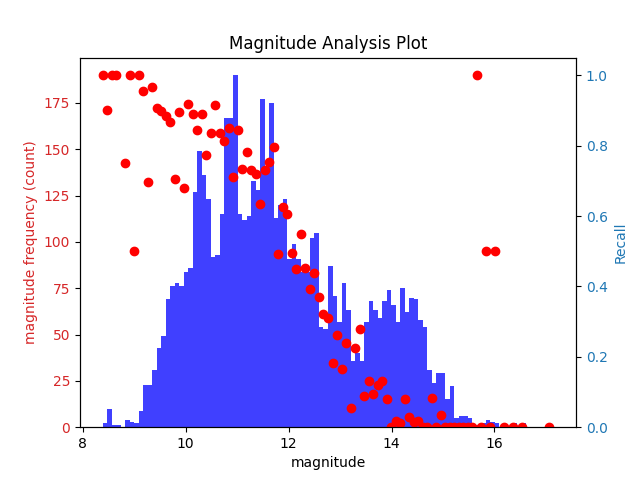}
    \caption{\textbf{Recall vs.\ visual magnitude of network trained on \textit{generated} images.}}
    \label{fig:mag-recall-gen}
\end{figure}

We further studied the relationship between visual magnitude ($m_V$) and network recall performance against the SatNet (real image) validation dataset. Visual magnitude is a measure of the apparent brightness of a target at an observation site. It is a logarithmic scale, with larger numbers indicating dimmer targets, and a difference of $5 m_V$ corresponding to a $100\times$ difference in apparent brightness. It is expected that dimmer targets will be harder to detect, and in the SatNet dataset, fewer dim targets are available to train and test on than brighter targets. Figures~\ref{fig:mag-recall-sim} and~\ref{fig:mag-recall-gen} show the recall of a network trained on simulated data and generated data respectively against the visual magnitude of SatNet validation objects. The distribution of validation object $m_V$ is shown in the background for reference. Although the two figures are similar, upon inspection it is clear that the network trained on generated data is performing slightly better on bright targets, and pushes the curve out further into better performance on critical dimmer targets. This implies that the generator is creating data more similar to the target domain from which the network is able to better learn the target task, and the critical task of detecting dim targets.

\section{Conclusions}
%%%%%%%%%%%%%%%%%%%%%%%%%%%%%%%%%%%%%%
\label{sec:conclusions_and_future}

A novel meta-architecture for self-attending task generative adversarial networks is shown to produce visually compelling fake images for a data domain transfer problem for space domain awareness. These images are quantitatively shown to approximate a target domain through cross-domain training of a detector network.

Future work may seek to explore SATGAN as a general augmentation approach across high-contrast imagery domains, e.g., microscopy. Additionally,  maturation of the evaluation metrics to include augmentation performance over additional target sensors, and in the presence of some synthetic noise may be useful. More extensive hyperparameter tuning may serve to further improve generalization to the target domain and provide evidence of the effect of various task networks on model performance.

%%%%%%%%%%%%%%%%%%%%%%%%%%%%%%%%%%%%%%%%%%%%%%%%%%%%%%%%%%%%%%%%%%%%%%%%%%%%%%%%%%%%%%%%%%%%%%%%%%%%%%
% \acknowledgments
% Project sponsors may be acknowledged in this section.

%%%%%%%%%%%%%%%%%%%%%%%%%%%%%%%%%%%%%%%%%%%%%%%%%%%%%%%%%%%%%%%%%%%%%%%%%%%%%%%%%%%%%%%%%%%%%%%%%%%%%%
\bibliographystyle{IEEEtran}
\bibliography{SatGAN}
% \begin{thebibliography}{1}
% 
% \bibitem{ITAR}
% U.S. Munitions List, Sections 38 and 47(7) of the Arms Export Control Act (22 U.S.C 2778 and 2794(7).
% 
% \bibitem{AeroConf}
% Aerospace Conference Web site: \underline{www.aeroconf.org}.
% 
% \end{thebibliography}

%%%%%%%%%%%%%%%%%%%%%%%%%%%%%%%%%%%%%%%%%%%%%%%%%%%%%%%%%%%%%%%%%%%%%%%%%%%%%%%%%%%%%%%%%%%%%%%%%%%%%%
\newpage
\thebiography
%% This biostyle allows you to insert your photo size 1in X 1.25in
\begin{biographywithpic}
{Nathan Toner}{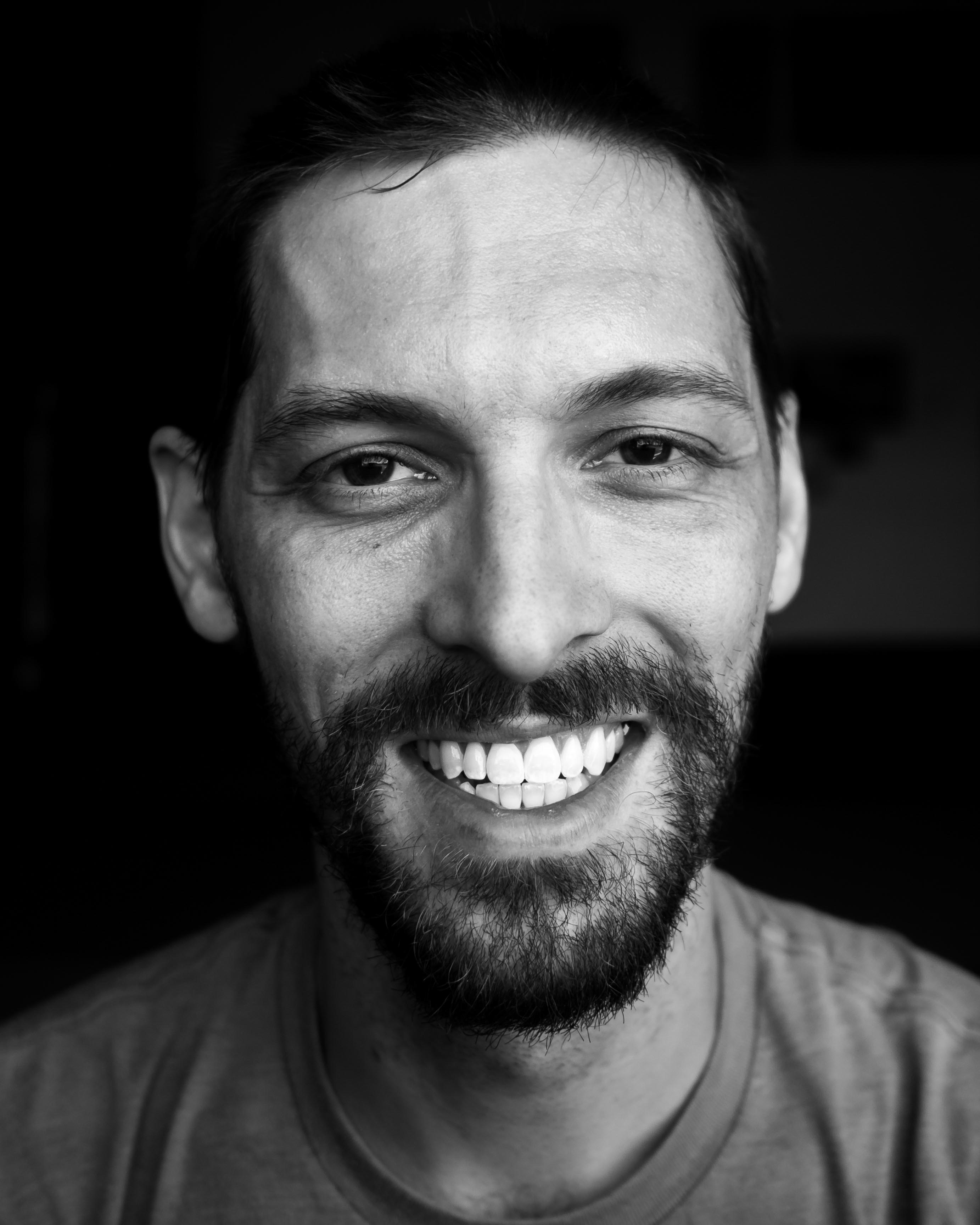}
received his Ph.D.\ in intelligent controls from Purdue University, West Lafayette, IN in 2016. Dr.\ Toner has been applying machine learning and machine vision algorithms to space domain awareness (SDA) problems since 2018, first with L3Harris and more recently with Centauri Corp.\ and KBR. Prior to SDA applications, Dr.\ Toner worked on close-in weapon and missile defense systems for the United States Navy. His current research interests include efficient networks for deployment to low size, weight, and power hardware and few- and one-shot learning.
\end{biographywithpic} 

\begin{biographywithpic}
{Justin Fletcher}{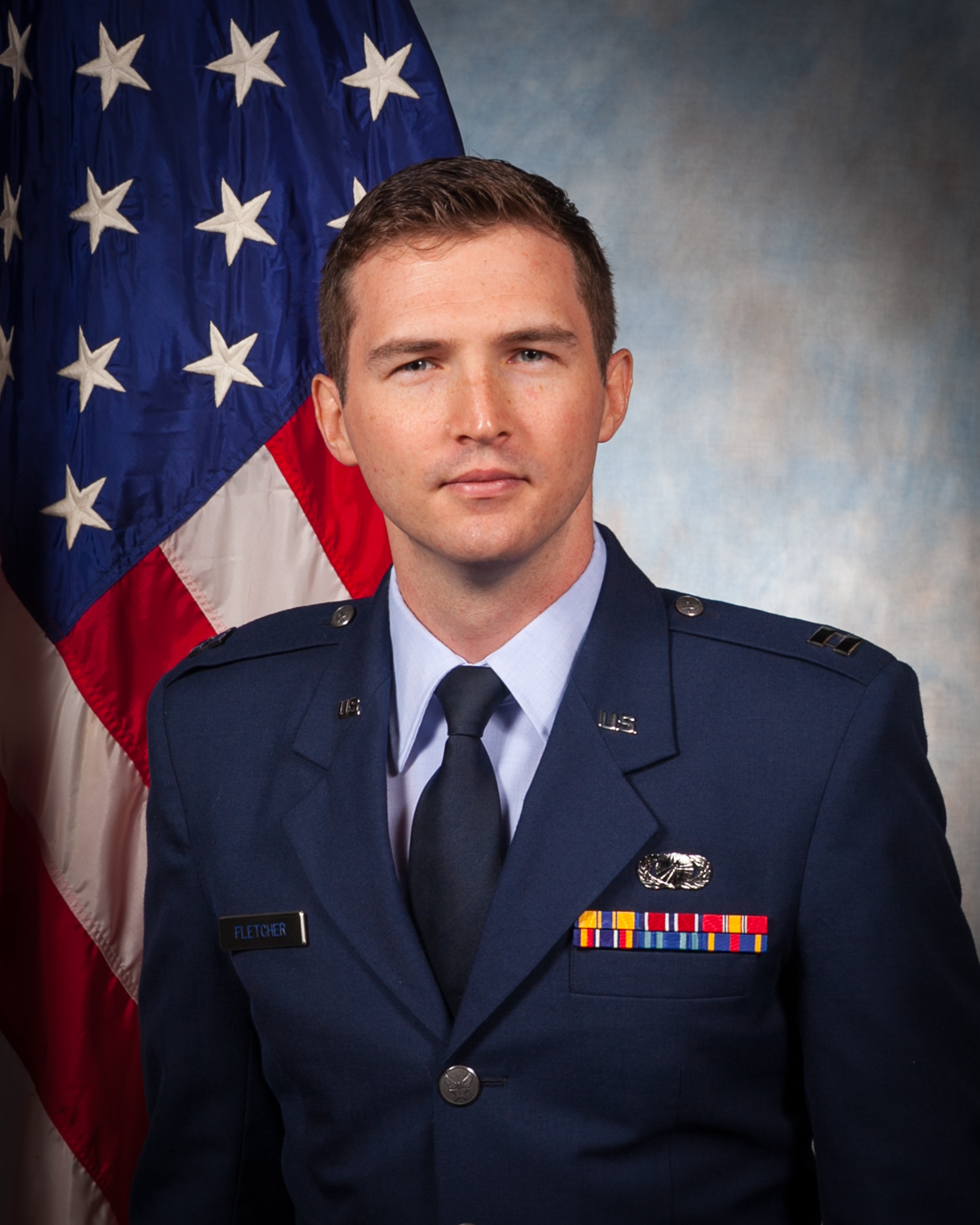}
received his M.S.\ degree in computer science from the Air Force Institute of Technology, Dayton, OH, USA, in 2016, with a concentration in machine learning. Mr.\ Fletcher is an advisor and subject matter expert on the enterprise adoption of deep learning technologies for the United States Space Force Space Systems Command and serves as a Major in the United States Air Force reserves. His current research interests include applications of computer vision to space object sensing, reinforcement learning for optical system actuation, and multi-agent autonomy.

\end{biographywithpic}

\end{document}